# Computational complexity reduction for BN2O networks using similarity of states

**Alexander V. Kozlov**
Department of Applied Physics
Stanford University
Stanford, CA 94305
phone: *(415) 725 - 8814*
e-mail: *alexvk@cs.stanford.edu*

**Jaswinder Pal Singh**
Department of Computer Science
Princeton University
Princeton, NJ 08544
phone: *(609) 258 - 5329*
e-mail: *jps@cs.princeton.edu*

## Abstract

Although probabilistic inference in a general Bayesian belief network is an *NP*-hard problem, computation time for inference can be reduced in most practical cases by exploiting domain knowledge and by making approximations in the knowledge representation. In this paper we introduce the property of similarity of states and a new method for approximate knowledge representation and inference which is based on this property. We define two or more states of a node to be similar when the ratio of their probabilities, the likelihood ratio, does not depend on the instantiations of the other nodes in the network. We show that the similarity of states exposes redundancies in the joint probability distribution which can be exploited to reduce the computation time of probabilistic inference in networks with multiple similar states, and that the computational complexity in the networks with exponentially many similar states might be polynomial. We demonstrate our ideas on the example of a BN2O network—a two layer network often used in diagnostic problems—by reducing it to a very close network with multiple similar states. We show that the answers to practical queries converge very fast to the answers obtained with the original network. The maximum error is as low as 5% for models that require only 10% of the computation time needed by the original BN2O model.

## 1   INTRODUCTION

A Bayesian belief network is a directed acyclic graph (DAG) whose nodes represent random variables and whose edges represent dependencies between the random variables. Belief networks are used for knowledge representation in diagnostic and forecasting software systems. Belief networks allow the user to answer *queries* about the probabilities of the states of one or several nodes, called *query* nodes, conditioned on other nodes, called *evidence* nodes. The process of finding these conditioned probabilities is called *probabilistic inference*.

Probabilistic inference is *NP*-hard for a general network with an arbitrary structure [Cooper, 1990]. Furthermore, even approximating inference in a general belief network is *NP*-hard [Dagum and Luby, 1993]. However, knowledge of the problem domain can help to reduce computation time of probabilistic inference. For example, Heckerman showed that probabilistic inference is linear in two-layer networks with noisy-OR interaction between nodes (BN2O networks) for negative evidence about nodes [Heckerman, 1989], and Heckerman and Breese showed that the noisy-OR interaction between nodes can be further simplified to reduce the number of parents per node which reduces computational complexity for networks with special structure [Heckerman and Breese, 1994]. Thus, the computational complexity of probabilistic inference can be managed in special cases.

In this paper we propose a new way of simplifying probabilistic inference in belief networks based on the property of *similarity of states*. Two or more states of a node are similar when the likelihood ratio does not depend on the instantiations of the other nodes in the network. The probability of one these states determines the probabilities of all states similar to it. If a model contains states that are almost similar, we can force the states to be similar and make probabilistic inference less computationally expensive. If we can make exponentially many states similar, the resulting computational complexity of probabilistic inference in the new model is polynomial in the size of the network. We call the new method *state space aggregation* since we explicitly aggregate states into groups to make probabilistic inference with them as with a group.

We demonstrate the new method on two examples of BN2O networks: One is a randomly generated BN2O network and the other is a BN2O network with noisy-OR coefficients randomly chosen from the practically important CPCS medical diagnostic network (the original CPCS network was constructed out of a Computer-based Patient Case Simulation database by R. Parker and R. Miller [Parker and Miller, 1987]).[1] We first convert a BN2O network to a cluster tree. A general form of the cluster tree for the BN2O network is a "Naive" Bayesian classifier with one large node representing all nodes in the first layer and many children representing nodes in the second layer. Although the resulting cluster node has exponentially many states, we can aggregate most of these states into a group of similar states. We then make probabilistic inference with these states as a group. We show that the resulting model provides a good estimate of the probabilistic inference results as compared to the original BN2O model and is better than the state abstraction method [Wellman and Liu, 1994] used for approximate inference.

The paper is organized as follows. In Section 2 we clarify the notations and conventions we use throughout the paper. In Section 3 we introduce BN2O networks and review the approaches to make probabilistic inference in them tractable. In Section 4 we define the property of similarity and develop our approach based on the modification of the original network to make a large subset of states similar. In Section 5 we demonstrate our ideas on the example of our randomly generated BN2O networks. In Section 6 we discuss the relation of the current technique to previous work. Finally, we conclude in Section 7.

## 2 NOTATIONS

In this paper, we use a special notation for a set of simple nodes in the original network. Thus, while we denote an individual simple node by a small letter $x$, with a possible subscript when further distinction is required, we denote a set of nodes by a capital letter $X$, again with a possible subscript. We call a set of simple nodes a *cluster node* since we can always represent a set of nodes as a single node which takes values from an expanded set of values, a set obtained by taking a direct product of the sets of values for the original nodes. A superscript of $X$, if it appears, denotes a particular state of the cluster node (i.e. a particular combination of states of the simple nodes in the set). We denote the number of simple nodes in the cluster node as $N(X)$ and the size of the state space as $|X|$.

In the networks considered in this paper (BN2O networks) all individual nodes are binary, i.e. can take only two values *false* and *true*. For the binary nodes, we assume that if $x_i$ is in the state *false*, then the variable $x_i$ is zero, and if $x_i$ is in the state *true*, then the variable $x_i$ is one. We assume a short form $p(\bar{i})$ to denote $p(x_i = \textit{false})$ and $p(i)$ to denote $p(x_i = \textit{true})$. For the cluster node, we assume a short form $p(X_i^s)$ to denote $p(X_i = X_i^s)$. Before we introduce the similarity of states property, let us consider BN2O networks further.

## 3 BN2O NETWORKS

A BN2O network is a two-layer network consisting of binary nodes with a noisy-OR dependence between the nodes in the first and the second layers. Let us take a medical diagnostic network as an example. The nodes in the first layer are diseases and the nodes in the second layer are symptoms (called findings in the medical literature) in this network. The nodes in the first layer (diseases) have one or several children in the second layer (findings). The noisy-OR interaction between the diseases and findings describes a causal independence assumption, i.e. that the ability of any single disease to cause a given symptom does not depend on the presence of the other diseases [Pearl, 1988].

A noisy-OR dependence between a finding node $x_{f_i}$ and its $n$ parent disease nodes $x_{d_j}$ can be characterized by $(n+1)$ real numbers from the interval $[0, 1]$: a leak and $n$ coefficients. The leak, which we denote $Leak(f_i)$, is the probability of the finding node in the absence of any of the diseases described by the network. A coefficient, which we denote $c_{ij}$, describes the ability of a disease $d_j$ to cause an increase in the probability $p(f_i)$ of the finding $f_i$. More precisely, the probability of the finding being absent $p(\bar{f_i})$ is multiplied by $(1 - c_{ij})$ each time a parent $x_{d_j}$ of the node $x_{f_i}$ changes its state from *false* to *true*. We can write the noisy-OR conditional probability in a closed form:

$$p(\bar{f_i}|x_{d_1}, x_{d_2}, \ldots, x_{d_n}) = \quad (1)$$
$$[1 - Leak(f_i)] \times \prod_j [1 - c_{ij} x_{d_j}];$$

$$p(f_i|x_{d_1}, x_{d_2}, \ldots, x_{d_n}) = \quad (2)$$
$$1 - [1 - Leak(f_i)] \times \prod_j [1 - c_{ij} x_{d_j}],$$

where we used our convention that $x_{d_j}$ is zero if the node $x_{d_j}$ is in the state *false* and one if it is in the state *true*. Trivially, if $c_{ij}$ is zero, the state of the parent does not affect the probability of the child, and if $c_{ij}$ is one, the *true* state of the parent forces the child to be *true* with probability one. An extension of the noisy-OR interaction to multiply-valued nodes is possible and is called noisy-MAX [Pradhan et al., 1994]. In the noisy-MAX interaction relations identical to (1) and (2) hold for the combined probability of the first $k$ states of the child node.

BN2O networks have been an object of study for a long time due to their potential applicability and due to the existence of a compact form for the noisy-OR conditional probabilities. Knowledge acquisition for BN2O network is also simplified since an expert has to assess only a small number of parameters—linear in the number of parents—to completely characterize the dependence. Also, probabilistic inference with BN2O networks has polynomial complexity in some special cases. For example, for negative findings the nodes in the first layer remain conditionally independent and the disease probabilities can be obtained by the summation of the probability distribution for the disease nodes:

$$p(x_{d_1}, x_{d_2}, \ldots, x_{d_n}) = \prod_i [1 - Leak(f_i)] \times \prod_j [1 - c_{ij} x_{d_j}] p(x_{d_j}), \quad (3)$$

where the first product is over the negatively instantiated findings. The computational complexity of the probabilistic inference with negative evidence is linear in the size of the network since the above probability distribution (3) is easily decomposable into factors of the form $[1 - c_{ik} x_{d_k}] p(x_{d_k})$. The new probabilities of the disease $d_k$ after instantiation are:

$$p^{(1)}(\overline{d_k}) = \frac{p(\overline{d_k})}{p(\overline{d_k}) + \prod_i [1 - c_{ik}] p(d_k)};$$
$$p^{(1)}(d_k) = \frac{\prod_i [1 - c_{ik}] p(d_k)}{p(\overline{d_k}) + \prod_i [1 - c_{ik}] p(d_k)}.$$

---
[1] The CPCS network is actually not quite a two-layer network. While several BN2O networks are similar to CPCS and are used in practice, they are proprietary and were not accessible for this paper.

Inference for positive evidence about nodes is more involved. We have to evaluate the sums over:

$$p(x_{d_1}, x_{d_2}, \ldots, x_{d_n}) = \prod_i \left(1 - [1 - Leak(f_i)] \times \prod_j [1 - c_{ij} x_{d_j}] p(x_{d_j})\right),$$

where the first product is over the positively instantiated findings. The probability of each of the diseases cannot be taken out of the summation easily for this sum and we have to evaluate it by expanding the outermost product in the above expression into a sum of products:

$$p(x_{d_1}, x_{d_2}, \ldots, x_{d_n}) = \sum_{X_E} (-1)^{\sum x_{f_i}} \Big[ \prod_{i: x_{f_i}=1} [1 - Leak(f_i)] \times \prod_j [1 - c_{ij} x_{d_j}] p(x_{d_j}) \Big], \quad (4)$$

where we sum over all possible instantiation of the cluster node $X_E$ consisting of the evidence nodes. Each of the terms in the sum (4) has structure identical to (3) and its contribution to the disease probabilities can be computed in linear time. However, the total number of terms in the sum is exponential in the number of evidence nodes, and the computational complexity of probabilistic inference with positive evidence is exponential in the number of evidence nodes [Heckerman, 1989].

The same complexity results can be obtained by considering topological transformations of networks with noisy-OR interactions [Heckerman and Breese, 1994]. The transformation can reduce the number of parents for a node in a network in special cases and is computationally equivalent to the above decomposition of the joint probability sums (4).

Besides these simplifications, probabilistic inference in a general BN2O network (as well as in the practical QMR-DT network [D'Ambrosio, 1994]) is computationally intractable for general positive evidence. In the practical QMR-DT network, users have to apply heuristic search or stochastic simulation methods to obtain approximate results. The above methods are unpredictable and sometimes fail to produce a satisfactory error bound on the result in time critical situations. In this case, we need to simplify the original model so that it deterministically produces a satisfactory answer in a known fixed amount of time. Let us now introduce similarity of states which we will use later to construct a model with polynomial complexity that is very close to the BN2O model.

## 4  SIMILARITY OF STATES

In this section, we introduce the *similarity* of states property, which we use later for our model reduction. The similarity of states is a form of independence in the network. It exposes a redundancy in the joint probability distribution and therefore can be used to make probabilistic inference faster.

**Definition 4.1:** The states $X_i^s$ and $X_i^{s'}$ of a node $X_i$ are *similar* with respect to a node $X_j$ if the ratio of the probabilities $p(X_i^s)/p(X_i^{s'})$ is invariant with respect to any instantiation of the node $X_j$:

$$\frac{p^{(k)}(X_i^s)}{p^{(k)}(X_i^{s'})} = \frac{p(X_i^s)}{p(X_i^{s'})} = const \quad (5)$$

for all instantiations of $X_j^k$ of the node $X_j$. We call the two states of a node simply *similar* if the two states are similar with respect to all the other nodes. ∎

The probability of one of the similar states determines the probabilities of the others. Let us take an example of car diagnosis. Given that a car doesn't start, the fact the the fuel tank is full increases the probability that the one of the spark plugs doesn't work. However, we can treat the probabilities that each one of the spark plugs failed as similar. Unless we look under the hood, the probability that one of the spark plugs has failed determines the probabilities of the failure of any other spark plug. The likelihood ratio of the spark failure probabilities stays the same.

This independence information was brilliantly used for the construction of similarity networks [Heckerman, 1990]. A similarity network is a construction consisting of a similarity graph and a collection of local knowledge maps corresponding to each edge in the similarity graph. Similarity networks were developed to simplify the *construction* of large and complex belief networks. They are a result of recognizing "specific forms of conditional independence" and developing special representations for them that simplify the knowledge acquisition. To build a similarity network, we first pick a distinguished node representing the hypotheses to be chosen from (for example, in a medical diagnosis problem, the hypotheses are the diseases). A node in a similarity graph is a hypothesis, and an edge indicates two hypotheses that are likely to be confused by an expert. For each such pair of hypotheses, we build a local knowledge map. A local knowledge map is a belief network for distinguishing between these two hypotheses. By focusing on constructing local knowledge maps, a person can concentrate on one manageable portion of the modeling task at a time.

Our goal, on the other hand, is to simplify probabilistic *inference* in complex belief networks. We do this by identifying redundancies in the joint probability distribution. The redundancy considered in this paper is the similarity of states and is related to the same "specific forms of conditional independence" as the similarity networks developed earlier. The local knowledge maps constructed by a knowledge engineer might, in fact, be used for identifying similar states. A local knowledge map would contain all nodes with respect to which the pair of hypotheses is not similar. In our definition, the concept of similarity of states is more general and can be applied to any node in the network, not just the distinguished node representing the hypotheses to be chosen from. We demonstrate how this concept can be applied for model reduction in the case of BN2O networks.

The following theorem shows the redundancy in the joint probability distribution introduced by the similarity of states:

**Theorem 4.2:** *The states $X_i^s$ and $X_i^{s'}$ of a node $X_i$ are similar with respect to a node $X_j$ iff the columns of the conditional probability matrix $(\mathbf{P}_{ji})_{qp} = p(X_j^q|X_i^p)$ corresponding to these two states $s$ and $s'$ are identical iff the columns of the probability distribution matrix $(\mathbf{D}_{ji})_{qp} = p(X_j^q, X_i^p)$ corresponding to the states $s$ and $s'$ are linearly dependent.*

The proof of the theorem is easy, and follows from the decomposition of the probability distribution for the two nodes into a product $(\mathbf{D}_{ji})_{qp} = p(X_j^q|X_i^p)p(X_i^p)$. In this form, any positive instantiation of the node $X_j$ in the state $X_j^k$ can be represented as removing all rows except $k$ from the matrix $\mathbf{D}_{ji}$. After the remaining row is normalized, the probability $p(X_i^p)$ can be read off from this row. Any negative instantiation of the node $X_j$ in the state $X_j^k$ can be represented as removing the row $k$ from the matrix $\mathbf{D}_{ji}$. After the remaining probability distribution is normalized, the probability $p(X_i^p)$ can be obtained by summing all numbers in $p$-th column. The proof is obtained by considering all possible instantiations of $X_j$.

Thus, the similarity of states uncovers a redundancy in the joint probability distribution. In linear algebra terms, two or several columns of the matrix representing the joint probability distribution are linearly dependent if the corresponding states are similar. If the columns are close to linearly dependent, we can approximate the joint probability distribution to make the states similar in order to simplify probabilistic inference. The theorem shows how we can introduce the similarity of states via conditional probabilities. We aggregate some of the states with almost identical conditional probability matrix columns and force them to be similar by assigning the same column to every one of these states.

Although for a general joint probability distribution the computational complexity of probabilistic inference is linear in the total number of states per any given node, the computational complexity of probabilistic inference with similar states is linear only in the total number of states that are not similar. By constructing models with exponentially many similar states we can reduce computational complexity from exponential to polynomial in some networks. In the next section we show that the precision of the reduced model as compared to the original model can be quite satisfactory in these cases.

## 5 EXAMPLE OF STATE SPACE AGGREGATION

We demonstrate the application of our state space aggregation method on the example of BN2O networks. We assume that a BN2O network has $n_1$ binary nodes in the first layer and $n_2$ binary nodes in the second layer, and that every node $x_{d_i}$ in the first layer is connected to every node $x_{f_j}$ in the second layer.[2] First, we describe our procedure and then compare the results for our reduced model to the results of a full BN2O network.

### 5.1 FORMALISM

We proceed by combining all nodes in the first layer into one large cluster node $X_D$ representing all possible diseases and their combinations. Node $X_D$ has an exponential number of states $2^{n_1}$, and we hope that some of these states can be made similar. We therefore partition the $2^{n_1}$ states into two subsets: One is the subset of $N_b$ *base* states, which we denote as $S_b$, and the other is the subset of $N_\sigma = |X_D| - N_b$ similar states, which we denote as $S_\sigma$. We will exploit different strategies for choosing the subset of states which we force to be similar (see subsections 5.2 and 5.3).

According to the definition of similar states (5), the contribution to the disease probabilities from the set of similar states $S_\sigma$ is a constant factor times the combined probability of the similar states $p(X_D^\sigma)$. The posterior probability of a disease is then computed as:

$$p^{(1)}(d_i) = \sum_{X_D^s \in S_b} p^{(1)}(X_D^s)x_{d_i} + \alpha(d_i)p^{(1)}(X_D^\sigma),$$

where the first term is the contribution to the disease probability from the base states and the second is the contribution to the disease probability form the similar states. The coefficients $\alpha(d_i)$ can be obtained from the prior probabilities:

$$\alpha(d_i) = \frac{p(d_i) - \sum_{s \in S_b} p(X_D^s)x_{d_i}}{1 - \sum_{s \in S_b} p(X_D^s)},$$

and are computed in linear time given the conditionally independent first layer nodes in the original network.

---

[2]Although sparse interconnection reduces the applicability of this method compared to the methods based on topological decomposition [D'Ambrosio, 1994, Heckerman and Breese, 1994], we will show that state space aggregation produces satisfactory results even for sparse BN2O networks. A combination of the topological method and methods based on state space aggregation is definitely possible but not considered in this paper.

The conditional probabilities for the base states match the conditional probabilities of the corresponding states in the original BN2O model. The conditional probability for the similar states—which has to be identical for every similar state—is chosen to preserve the prior probabilities of the findings:

$$p(f_j|X_D^\sigma) = \frac{p(f_j) - \sum_{s \in S_b} p(f_j|X_D^s)p(X_D^s)}{1 - \sum_{s \in S_b} p(X_D^s)}. \quad (6)$$

The last equation is an application of the Bayes' rule for a finding node $x_{f_j}$ and the aggregated similar state.

As we can see, the computation to transform the model to a reduced model involves simple summations over the base states. The computational requirements in this state aggregation model are thus the same as in the state space abstraction model [Wellman and Liu, 1994] in which the answer to a query is inferred by summing over the base states only and completely ignoring the rest of the states. Our model accounts for some of the ignored probability mass via the coefficients $\alpha(d_i)$. Let us now see how the state space aggregation model helps to increase the precision of probabilistic inference in BN2O networks.

## 5.2 RANDOMLY GENERATED BN2O NETWORKS

To demonstrate how the state aggregation model can help improve the precision of the model and reduce the computation time of probabilistic inference, we first generated a BN2O network with randomly chosen noisy-OR coefficients drawn from a $Beta(2,4)$ distribution (with the expected value $\langle c_{ij} \rangle = 1/3$). If a large number of nodes in the first layer are in the state *true*, we expect that the probability of any finding is close to one. Thus, we make similar all states of the cluster node $X_D$ in which the number of diseases present is larger than some fixed number $d_{\max}$:

$$X_D^s \in S_\sigma \quad \text{if} \quad \sum_{i=0}^{N(X_D)} x_{d_i} \geq d_{\max}.$$

The number of base states in this case is polynomial in the number of the first layer nodes:

$$N_b = \sum_{i=0}^{d_{\max}} \binom{n_1}{i} \leq n_1^{d_{\max}},$$

for $d_{\max} > 1$, and the reduction of the original BN2O model to the model with the aggregated similar states has polynomial complexity.

First, we analyzed the maximum absolute error for the queries about the probability of each of the diseases (nodes in the first layer) given different instantiations with different number of positive findings (nodes in the second layer). The results of the simulations for the BN2O network consisting of 18 nodes in the first layer and 18 nodes in the second layer are presented in Figures 1 and 2 (for the state space abstraction and the state space aggregation models respectively). The error in the state space aggregation model is much smaller (about an order of magnitude for high $d_{\max}$) than the error in the state space abstraction model, where the set $S_\sigma$ is completely disregarded. Also, the error for the instantiations with the large number of findings present—the region where the probabilistic inference is computationally very expensive—is almost independent of the number of diseases present. Since the errors introduced by the instantiations of different findings can be considered independent, we expect the maximum error to be $O(\sqrt{N(X_E)})$, i.e. to grow as a square root of the number of instantiated nodes. For our network, the maximum absolute error is less than 0.005 for $d_{\max} > 5$ over all possible instantiations of the nodes in the second layer.

Second, we analyzed the behavior of the maximum relative error in the above network. The relative error as opposed to the absolute errors might be more important for some problems. For example, the probability of a life-threatening disease being $10^{-3}$ is substantially better than the probability of it being $10^{-2}$, and the relative error of 10 shows this more clearly than the absolute error of 0.099. Figure 3 shows the maximum relative errors for the above models over all possible instantiations of the second layer nodes. The error in the state space aggregation method is about an order of magnitude lower than in the state space abstraction method for high $d_{\max}$. Our method gives superior precision as it partially accounts for the states completely ignored in the state space abstraction method. The error decreases as the combined prior probability of the similar states (i.e. before any instantiation), which is shown on the same plot. The maximum relative error is less than 0.01 for $d_{\max} > 6$ over all possible instantiations of the nodes in the second layer.

## 5.3 CPCS-LIKE NETWORKS

Although the above generated networks do not have the structure that real practical networks have, the state space aggregation method can be extended to practical problems given some insight into the problem domain. For example, the rule for selecting the base states above can be formulated in the domain language: If the number of diseases present is large, then with a high probability a patient has any imaginable finding present. Thus, the states are almost similar already and we do not change the conditional probabilities much by forcing them to be similar. We can also argue that the cases with more than a certain number of diseases present occur rarely in practice and are not important for diagnosis (have a low combined probability mass). Although the validity of these specific rules might be arguable in the medical domain, rules of these type can definitely lead to state space aggregation and simplification of probabilistic inference.

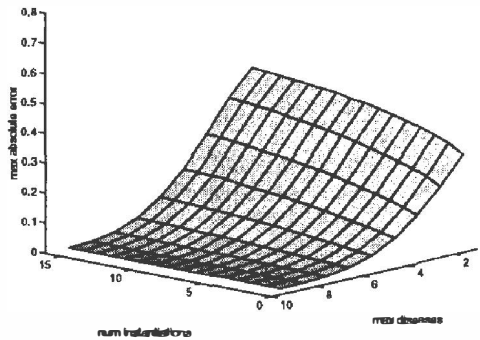

Figure 1: The maximum absolute error of the answer to a query about a disease probability for the state space abstraction method adopted from [Wellman and Liu, 1994]. The maximum error was found by an exhaustive search over all possible positive finding instantiations. The computational complexity of probabilistic inference is $O(n_1^{d_{max}})$.

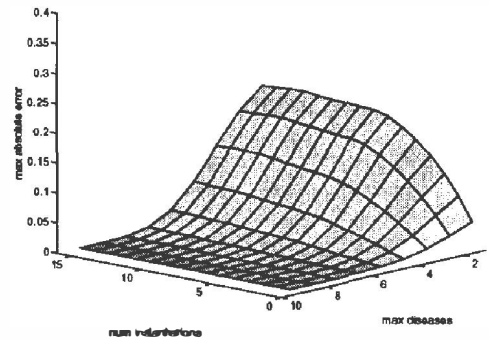

Figure 2: The maximum absolute error of the answer to a query about a disease probability for the reduction based on state space aggregation. The maximum error was found by an exhaustive search over all possible positive finding instantiations. The computational complexity of probabilistic inference is still $O(n_1^{d_{max}})$ as in the state space abstraction method.

The similarity of states is present already in many practical networks. For instance, Figure 4 shows that the majority of the noisy-OR coefficients in the CPCS network are concentrated close to round numbers like 0, 0.2, 0.5, 0.8, or 1 since further precision is not necessary for the diagnostic problem at hand. Besides, our study of the noisy-OR distributions in this network show that in many cases the coefficients are equal exactly, implying the appropriate redundancy in the joint probability distribution. Identification of these similar states, however, is best done by a domain expert.

To study the effect of structure on the precision of the state space aggregation and to demonstrate another rule for choosing the set of base states, we built a BN2O network with coefficients drawn randomly from the set of real noisy-OR coefficients in the CPCS network. The presence of the noisy-OR coefficients that are close to zero or one—which constitute about 50% of the total number of noisy-OR coefficients in the CPCS network—makes the state aggregation more complex and requires a better algorithm for the selection of similar states. Consider the states with the number of diseases present equal to $d_{max}$ as in the previous subsection. A subset of $d_{max}$ diseases might no longer cause a finding if the coefficients for this subset are close to zero. The conditional probability of the finding is no longer close to one, and including this state in the set of similar states and altering the corresponding conditional probability of the finding can affect the accuracy of the network.

Table 1: Errors in the CPCS-like BN2O network

| $\lambda$ | $N_b/|X_D|$ | max abs error | max rel error |
|---|---|---|---|
| 0.3 | 4.5% | $4.4 \times 10^{-2}$ | $1.8 \times 10^{0}$ |
| 0.4 | 10.6% | $6.2 \times 10^{-3}$ | $5.4 \times 10^{-2}$ |
| 0.5 | 17.4% | $5.8 \times 10^{-4}$ | $1.0 \times 10^{-2}$ |
| 0.6 | 27.6% | $1.3 \times 10^{-4}$ | $1.7 \times 10^{-3}$ |

To cope with this situation, we had to modify the base state selection algorithm for the CPCS-like BN2O network. We consider a state of the cluster node $X_D$ to be a base if the conditional probability of any finding given this state is bigger than a fixed parameter $\lambda$. The results for this base state selection policy are given in Table 1. For a relative error of 5% we need to account exactly for only 10% of the total number of states, thus reducing the computation time of the diagnosis ten times.

These simple examples show that a large state space of a node can be managed by having many similar states in practical problems, and thus the large sizes of the cliques in the join tree can be managed by introducing similarity between states. Given that the state spaces of the join tree nodes can be very large, we are likely to find exponentially many states that can be aggregated

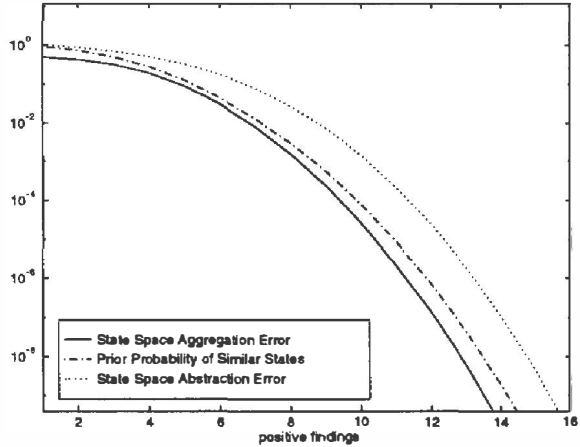

Figure 3: The combined prior probability of the similar states $p(X_D^{\sigma})$ and the maximum relative errors $|\Delta p^{(1)}(d_i)/p^{(1)}(d_i)|$ of the posterior disease probabilities over all possible queries as a function of $d_{\max}$. All three curves have the same asymptotic behavior. The error in the state space aggregation method is smaller since it partially accounts for the probability mass that is completely ignored in the state space abstraction method.

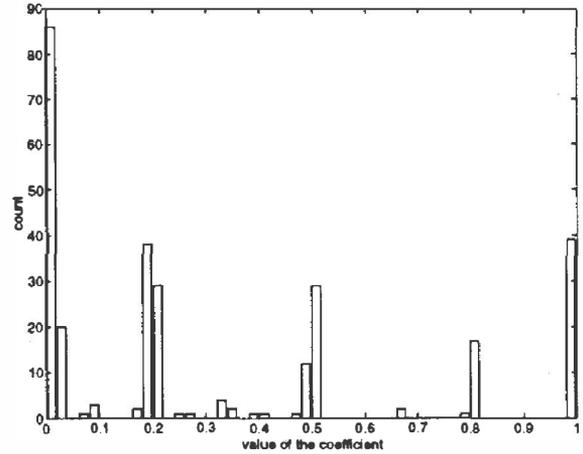

Figure 4: The distribution of the noisy-OR coefficients in the practical CPCS network. Most coefficients are close to round numbers (0, 0.2, 0.5, 0.8, or 1). If a coefficient is close to zero, the state of the parent node only slightly affects the probability of the child. If a coefficient is close to one, the *true* state of the parent causes the child to be in the state *true* also.

into groups, especially if we have some insights into the underlying problem.

## 6 RELATION TO PREVIOUS WORK

Since the BN2O networks are practically important, a few approximate algorithms has been developed distinctively for this type of networks. The Quickscore uses the noisy-OR properties described in Section 3 to rearrange the summation of the joint probability distribution [Heckerman, 1989], making the probabilistic inference exponential in the number of positively instantiated nodes, not the number of nodes in the first layer as given by the direct triangulation. The TopN algorithm [Henrion, 1991] tries to bound the (ratios of) posterior probabilities for the most likely $N$ diseases by searching in a subspace of the full probability distribution for the first layer nodes. Stochastic simulation methods [Henrion, 1988] have been specifically extended to sample the joint probability distribution of BN2O networks. The approach taken in this paper differs from the previous ones in that we reduce the complexity of probabilistic inference by making approximations in the knowledge representation, not by making approximations in the inference procedure. The reduced and full models take the same amount of space for their representation (the number of coefficients to completely specify the dependence is exactly the same), but the reduced model produces results of almost the same quality in polynomial amount of time. On the other hand, our approach is close in spirit to the previously developed TopN and state space abstraction algorithms in that it tries to account for the major probability mass of the joint probability distribution exactly, while making approximations about the rest of the probability mass.

Our method is directly related to the proposed earlier general approach to complexity reduction using sensitivities instead of conditional probabilities. [Kozlov and Singh, 1995], and in fact was first derived in terms of sensitivities. In the previous work we suggested reducing the computational complexity of probabilistic inference for general networks by reducing the rank of the sensitivity matrices by averaging out the columns of the sensitivity matrix. It can be shown that assigning the same value to conditional probabilities without changing the prior probabilities of nodes is equivalent to averaging out sensitivity matrix elements over a subset of states. In the case of BN2O networks this averaging is reduced to identifying the similar subset of the cluster node $X_D$ and assigning the same conditional probability to all these states. However, the methods based on sensitivities are likely to result in a larger class of complexity reduction methods, particularly for multiply-valued nodes where the analysis in terms of traditional conditional probabilities is complicated.

## 7 SUMMARY AND FUTURE WORK

We define the property of similarity of states and use it for model reduction. Two states of a node are similar if the ratio between the probabilities of the two states remains constant after any instantiation of other nodes in the network. We show that the similarity of states property can be exploited to perform probabilistic inference more efficiently. The computational complexity of probabilistic inference in networks with similar states is determined by the total number of non-similar states instead of the total number of states, and might be polynomial in the size of the network if exponentially many states are similar.

We show a relation between the similarity of states property and the redundancies in the joint probability distribution. The states are similar if and only if the corresponding columns in the joint probability distribution are linearly dependent. We find a generic way of identifying similarity of states and enforcing the similarity property on states that we want to make similar through conditional probabilities. Thus, we can reduce computation time of probabilistic inference by enforcing the similarity of states in a model. The accuracy of the reduced model is determined by how similar the states are in the original problem already. We show that the BN2O models can be readily reduced to a model with exponentially many similar states, and that the reduced model produces results very close to the original model for all queries of practical importance.

The proposed method of complexity reduction is related to the developed earlier TopN [Henrion, 1991] and the state space abstraction [Wellman and Liu, 1994] methods. As in the above methods, we also try to account for the major probability mass in the joint probability distribution exactly, but make some approximations about the unaccounted-for probability mass. When the accounted-for probability mass is substantial, all methods produce almost exact results. However, our method produces superior accuracy as it estimates the contribution from the rest of the probability mass and performs better on real networks.

The model reduction described in this paper can be readily expanded to any other network represented as a cluster tree (a singly-connected Markov network of cluster nodes). The cluster nodes will have exponentially many states and many of these states are likely to be almost similar. The method can readily be extended by building several groups of similar states per cluster node, thus improving the accuracy without much computation overhead. In this paper we have shown a successful application on two BN2O networks: One randomly generated and the other build based on the CPCS medical diagnostic network. For the network we studied, the error can be as little as 5% for the reduced problem while requiring only 10% of the computation time needed by the original problem. Further applications of the new approach are of course necessary, and we are actively pursuing the application to practical belief networks and expert systems.

### Acknowledgments

We thank Randy Miller and University of Pittsburgh for supplying the CPCS network. We also thank Daphne Koller, Malcolm Pradhan, and Lise Getoor for reading the manuscript and valuable comments, John Hennessy for his support and guidance, and ARPA for financial support under contract no. N00039-91-C-0138.